\documentclass[runningheads]{llncs}
\usepackage{graphicx}
\usepackage{inconsolata}
\usepackage{graphicx}
\usepackage{hyperref}
\usepackage{url}
\usepackage{times}
\usepackage{epsfig}
\usepackage{graphicx}
\usepackage{bigstrut}
\usepackage{multirow}
\usepackage{xcolor}
\usepackage{booktabs,multirow}
\usepackage{xspace}
\usepackage{subcaption}

\newcommand{\sgraph}{word-relation graph\xspace}
\newcommand{\sgraphs}{word-relation graphs\xspace}

\newcommand{\ergraph}{entity-relation graph\xspace}
\newcommand{\ergraphs}{entity-relation graphs\xspace}

\newcommand{\system}[1]{\textsc{#1}\xspace}

\newcommand{\rel}[1]{\textit{#1}\xspace}

\newcommand{\languagexlm}{\system{LanguageXLM}}
\newcommand{\ourmodel}{\system{LaGNN}}

\newcommand{\proximatev}{\rel{proximate\_(V)}}
\newcommand{\proximateh}{\rel{proximate\_(H)}}
\newcommand{\qa}{\rel{question-answer}}
\newcommand{\hq}{\rel{header-question}}
\newcommand{\nl}{\rel{no-relation}}
\newcommand{\ww}{\rel{same-entity}}

\usepackage{amsthm,amsmath,amsfonts,bm,xspace}
\usepackage{color}

\def\vs{{\em v.s.}\xspace}

\def\eqref#1{(\ref{#1})}

\def\1{\bm{1}}

\def\rmA{{\mathbf{A}}}

\def\vb{{\bm{b}}}

\def\vd{{\bm{d}}}
\def\ve{{\bm{e}}}

\def\vh{{\bm{h}}}

\def\vs{{\bm{s}}}

\def\vu{{\bm{u}}}

\DeclareMathAlphabet{\mathsfit}{\encodingdefault}{\sfdefault}{m}{sl}
\SetMathAlphabet{\mathsfit}{bold}{\encodingdefault}{\sfdefault}{bx}{n}

\def\gE{{\mathcal{E}}}

\def\gG{{\mathcal{G}}}

\def\gN{{\mathcal{N}}}

\def\gV{{\mathcal{V}}}

\def\gZ{{\mathcal{Z}}}

\begin{document}
\title{Language Independent Neuro-Symbolic Semantic Parsing for Form Understanding}
%
%
\author{Bhanu Prakash Voutharoja \orcidID{0000-0002-1352-2858} \and
Lizhen Qu \thanks{Corresponding author} \orcidID{0000-0002-7764-431X} \and Fatemeh Shiri \orcidID{0000-0001-8752-2132}} 

\institute{Monash University, Clayton VIC 3800, Australia \\
\email{\textbf{voutharoja.bhanu06@gmail.com, \{lizhen.qu, fatemeh.shiri\}@monash.edu}}}


%
\maketitle              
\begin{abstract}
Recent works on form understanding mostly employ multimodal transformers or large-scale pre-trained language models. These models need ample data for pre-training. In contrast, humans can usually identify key-value pairings from a form only by looking at layouts, even if they don't comprehend the language used. No prior research has been conducted to investigate how helpful layout information alone is for form understanding. Hence, we propose a unique \ergraph parsing method for scanned forms called \ourmodel, a language-independent Graph Neural Network model. Our model parses a form into a \sgraph in order to identify entities and relations jointly and reduce the time complexity of inference. This graph is then transformed by deterministic rules into a fully connected \ergraph. Our model simply takes into account relative spacing between bounding boxes from layout information to facilitate easy transfer across languages. To further improve the performance of \ourmodel, and achieve isomorphism between \ergraphs and \sgraphs, we use integer linear programming (ILP) based inference. Code is publicly available at \href{https://github.com/Bhanu068/LAGNN}{https://github.com/Bhanu068/LAGNN}.

\keywords{Document Layout Analysis  \and Graph Neural Network \and Language Independent \and Deep Learning}
\end{abstract}
\section{Introduction}
\label{sec:intro}
\begin{figure}[t]
    \begin{center}
    \includegraphics[width=1.0\textwidth]{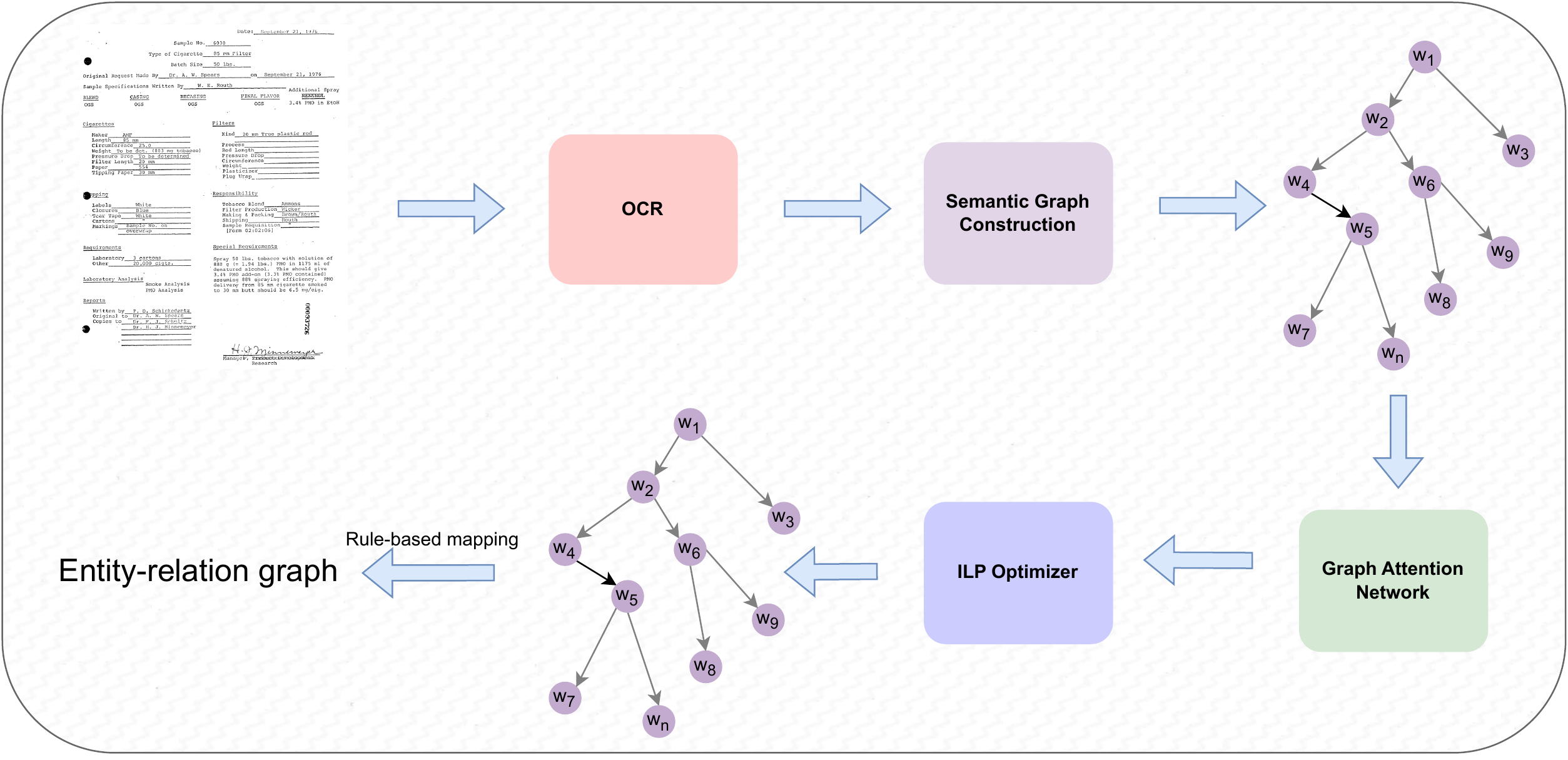}
    \end{center}
    \caption{Illustration of the pipeline of our proposed method.}
    \label{fig:form_intro}
\end{figure}

Despite the growing popularity of e-forms, paper forms are still widely used to collect data by various types of organizations, from government agencies to private companies. A large body of collected data, especially historical data, is still available only in paper forms or scanned document images. To digitize such data, we introduce the task of \textit{\ergraph parsing for form understanding}, which maps the document image of a form to a structured \ergraph. As a result, users can explore and analyze semantic information in such \ergraphs without any need to read the original document images. 

Entity relation graphs are introduced for forms in FUNSD \cite{funsd}, which are annotated on a small sample of scanned forms. Herein, an entity is a group of words representing a semantic and spatial standpoint, such as question and answer, and a relation is a directed edge between two entities, as illustrated in Fig. \ref{fig:form_intro}. Such graphs are layout-agnostic. However, such graphs are not always fully connected via those relations due to neglecting the layout information between entities. In contrast, form designers often put semantically relevant entities close to each other in a form, hence spatial proximateness is informative for semantic relevance between entities or relations. 

Prior models on form understanding tackle at least two subtasks in the sequel, which are entity recognition and relation extraction~\cite{layoutxlm}. The former identifies a group of words belonging to the same semantic entity and the label of the entity, while the latter predicts the relation between any two entities. Such a pipeline may easily lead to error propagation because the models for relation extraction are not able to fix errors of entity recognition, not to mention the exploitation of proximateness between relations to build a fully connected \ergraph. However, \ergraph parsing by tacking both subtasks jointly is challenging because the time complexity of inference is quadratic to the number of tokens in a form, as shown in Sec. \ref{sec:methodology}.

The use of large-scale pre-trained language models or multimodal transformers dominates in the recent studies on form understanding~\cite{layoutlm,layoutlmv3,layoutxlm}. However, such models require large-scale data for pre-training, especially for multilingual document understanding. For example, \languagexlm needs 30 million documents in 53 languages for pre-training~\cite{layoutxlm}. Adding any new languages or new document collections often requires re-training of the models. In contrast, humans are often capable of recognizing key-value pairs from a form by only using layout information, even though they may not understand the language in the form. However, no prior studies explore to what degree layout information is useful for language-agnostic form understanding.

    In this work, we propose a novel language-agnostic Graph Neural Networks (GNN) model for \ergraph parsing of scanned forms, coined \ourmodel. To facilitate navigation in an \ergraph and retain proximateness of entities and relations after parsing, we add two types of relations for proximateness to \ergraphs: i) vertically proximate, coined \proximatev, and ii) horizontally proximate, coined \proximateh. To mitigate error propagation in pipeline approaches, we introduce a \sgraph representation so that our model parses a document image into such a graph, which is subsequently converted to a fully connected \ergraph by deterministic rules. This simplification enables linear inference time. To support language-agnostic form understanding without using pre-trained language models, our model only considers features extracted from layout information. Furthermore, we apply integer linear programming (ILP) based inference to ensure that the generated graphs satisfy the properties of \ergraphs.   



The main contributions of this paper are summarised as follows:
\begin{itemize}
  \item We propose a language-agnostic GNN model, coined \ourmodel, for parsing scanned forms into a \sgraph, which is isomorphic to a fully connected \ergraph with two new relations based on proximateness.
  \item We propose a designated ILP-based inference method to ensure that i) the generated graphs are fully connected, and ii) relations in a graph are logically coherent.
  \item The extensive experimental results show that our model significantly outperforms the competitive baselines in terms of all metrics in the monolingual settings and the averaged metrics in the zero-shot multilingual settings.
\end{itemize}

\section{Related Work}
\label{sec:related_work}
Deep learning techniques have dominated document interpretation tasks \cite{Yang2017LearningTE,Augusto,Siegel2018ExtractingSF} in the past decade. Grid-based techniques~\cite{katti-etal-2018-chargrid,denk2019bertgrid,Lin} were suggested for representing 2D documents. In these techniques, first character-level or word-level embeddings are used to represent text, and later CNNs are used to categorize them into different field types. 

Self-supervised pre-training has had a lot of success lately. Recent work on structured document pre-training~\cite{layoutlm,layoutlmv2,layoutxlm,lilt,structurallm} has pushed the boundaries, drawing inspiration from the success of pre-trained language models on multiple downstream NLP tasks. The BERT architecture was altered by LayoutLM \cite{layoutlm} by including 2D spatial coordinate embeddings. By considering the visual aspects as independent tokens, LayoutLMv2 \cite{layoutlmv2} outperformed LayoutLM. To optimize the use of unlabeled document data, extra pre-training activities were investigated. In contrast to StructuralLM \cite{structurallm}, who suggested cell-level 2D position embeddings and the accompanying pre-training target, SelfDoc \cite{selfdoc} developed the contextualization across a block of text. To unify many issues surrounding natural language, TILT \cite{Tilt} suggests a pre-trained layout-aware multimodal encoder-decoder Transformer. The useful coarse-grained information like natural units and salient visual regions are ignored by the current layout-aware multimodal Transformers. In an effort to include coarse-grained information into pre-trained layout-aware multimodal Transformers, \cite{eernie} argues that both fine-grained and coarse-grained multimodal information is useful for document understanding and proposes a multi-grained and multimodal transformer, ERNIE-mmLayout. 

However, the preceding Structured Document Understanding (SDU) methods mostly rely on a single language, which is usually English, making them rather constrained in terms of multilingual application scenarios. LayoutXLM \cite{layoutxlm} was the first to incorporate a multilingual text model InfoXLM \cite{infoxlm} initialization to LayoutLMv2 framework for multilingual pre-training with structured documents. However, a laborious procedure of multilingual data collecting, cleansing, and pre-training was necessary. To address this issue, LiLT \cite{lilt}, a straightforward yet powerful language-independent layout Transformer for monolingual/multilingual structured document interpretation was proposed. LiLT employs BiACM to achieve language-independent cross-modality interaction and an efficient asynchronous optimization technique for both textual and non-textual flows in pre-training using two pre-training objectives.

Current state-of-the-art approaches to these document understanding challenges have made use of the power of large pre-trained language models, focusing on language more than the visual and geometrical information in a text, and end up using hundreds of millions of parameters in the process \cite{doc2graph}. Additionally, the majority of these models are trained using a massive transformer pipeline, which necessitates the pre-training of enormous amounts of data. In this sense, models that are independent of language were proposed \cite{7,Sarkar2020DocumentSE}. \cite{7} concentrated on identifying entity relationships in forms using a straightforward CNN as a text line detector, and then they find key-value relationship pairs using heuristics based on the model's scores  for each connection candidate. Later, \cite{Sarkar2020DocumentSE} reformulated the issue as a semantic segmentation (pixel labelling) task with a focus on extracting the form structure. They employed a U-Net based architecture pipeline, which was quite effective at concurrently predicting all levels of the document hierarchy. For form understanding, \cite{Carbonell} employed GCNs to solve the entity grouping, labelling, and entity linking tasks. They did not utilize any visual features and instead used word embeddings and bounding box information as the main node features, and k-nearest neighbours to obtain edge features. The FUDGE \cite{8} framework was then created as an extension of \cite{7} to help with form understanding. It proposes relationship pairings using the same detection CNN as in \cite{7}, considerably improving the state-of-the-art on both the semantic entity labeling and entity linking tasks. Then, because predicting key-value connection pairs and the semantic labels for text entities are two tasks that are closely associated, a GCN was implemented using plugged visual features from the CNN. Inspire by FUDGE \cite{8}, a task-agnostic GNN-based framework called Doc2Graph \cite{doc2graph} that adopts a similar joint prediction of both the tasks, semantic entity labeling and entity linking utilizing a node classification and edge classification module, respectively, without relying on heuristics to establish associations between words or entities was developed. To take advantage of the relative location of document objects via polar coordinates, a novel GNN architecture pipeline with node and edge aggregation functions is implemented.

\section{Methodology}
\label{sec:methodology}
\begin{figure}[t]
    \begin{center}
    \includegraphics[width=0.88\textwidth]{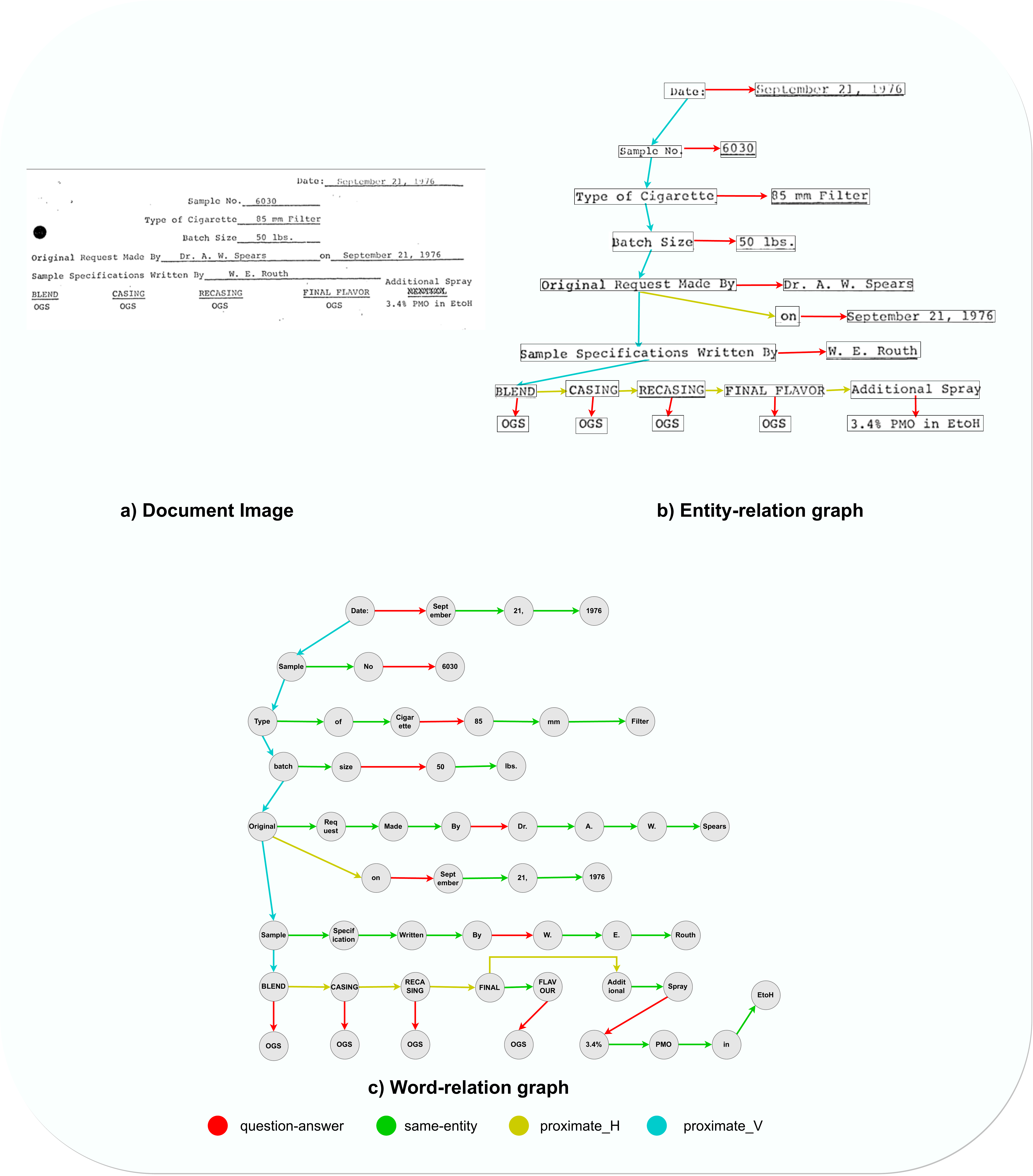}
    \end{center}
    \caption{Illustration of our proposed \ergraph and \sgraph.}
    \label{fig:word_relations}
\end{figure}


Entity-relation graph parsing for form understanding is concerned with mapping a document image to an \ergraph. To reduce inference time complexity, we propose to map a scanned form image to a novel \sgraph, which is isomorphic to the corresponding \ergraph. Then \ergraphs can be directly constructed from \sgraphs using deterministic rules.

Formally, an \ergraph is denoted by $\gG^e = \{\gV^e, \gE^e\}$, composed of a set of entities $\gV^e$ and a set of relations between entities $\gE^e$. Each entity is a word sequence $[w_0, ..., w_m]$ labeled by a type $z^e \in \gZ^e$, and each word is associated with a bounding box $(x_{left}, y_{top}, x_{right}, y_{bottom})$~\cite{funsd}. The set $\gZ^e$ includes \textit{question}, \textit{answer}, and \textit{section}. A relation $(v_a, y, v_b) \in \gE^e$ is denoted by a directed edge from an entity $v_a$ to another entity $v_b$ labeled by $z \in \gZ^r$, where $\gZ^r$ includes the following relations.  
\begin{itemize}
    \item \textit{Question-answer}, denoted by a directed edge from a question to the corresponding answer.

    \item \textit{Header-question}, denoted by a directed edge from a section and a question belonging to that section.

    \item \textit{Proximate\_(V)}, denoted by an undirected edge between an entity A and the first entity B in the first line below it, which is not an answer and there is no relation between A and B or the parent of B.

    \item \textit{Proximate\_(H)}, denoted by an undirected edge between an entity A and the closest entity labeled as question or header to its right in the same line. 
\end{itemize}

In contrast, a \sgraph $\gG^w = \{\gV^w, \gE^w\}$ depicted in Fig.~\ref{fig:word_relations} comprises a set of nodes $\gV^w$ consisting of only the words in a document, and a set of relations $\gE^w$ between those words. A relation in $\gE^w$ is a tuple $(w_a, z, w_b)$ from a word $w_a$ to another word $w_b$ labeled by $z \in \gZ^w$, where the edge label set $\gZ^w$ augments $\gZ^r$ by adding word-level relations. More details are provided in Sec.\ref{sec:word-relation_graph}

    
As illustrated in Fig.\ref{fig:form_intro}, to parse document images into \sgraphs, we first apply an off-the-shelf Optical Character Recogniser docTR \cite{doctr2021} to map a document image into a set of cells, whereby each cell is a sequence of bounding boxes and each bounding box corresponds to a word. Then we apply our model \ourmodel to estimate the probabilities of relations between each possible pair of words largely based on relative distances between bounding boxes. To construct a \sgraph, we apply a designated ILP inference algorithm to assign relation labels to word pairs \textit{jointly}. 

\subsection{Word-Relation Graph}
\label{sec:word-relation_graph}
A naive solution that directly parses a document image to an \ergraph results in a time complexity quadratic to the number of words in a document in the worst case. First, a model needs to identify if a word belongs to the same group of its adjacent words, followed by predicting relations between any pairs of word groups. In the worst case, if each entity comprises only one word, the time complexity is $O(n^2)$ because the number of edge predictions is $n \times n-1$, where $n$ is the number of words in a document. In fact, the average word lengths of entities in FUNSD \cite{funsd} is 3.4, the actual inference time complexity is not far from the worst case. The cost is estimated without considering the one for estimating word groups for each entity.

We map \ergraphs to \sgraphs so that \ergraph parsing becomes the task of predicting relations between words in a \sgraph. To eliminate the task of entity recognition, we introduce an undirected relation \ww to link two adjacent words in the same word sequence of an entity. For a relation $(v_a, z, v_b)$ between two entities in an \ergraph, we adapt the relation labels in \ergraphs to word-level relations:

\begin{itemize}
    \item \textit{Question-answer}, the last word in a question is linked to the first word in the corresponding answer.
    \item \textit{Header-question}, the first word in a section is linked to the first word of the corresponding question.
    \item \textit{Proximate\_(V)}, the first word in $v_a$ is linked to the first word in $v_b$.   
    \item \textit{Proximate\_(H)}, the last word in $v_a$ is linked to the first word in $v_b$.
    \item \textit{Same-entity}, the adjacent words within an entity are connected with this relation.
\end{itemize}

The conversion process is deterministic because i) for a relation between two entities, either the first or the last word of an entity is linked, and ii) \ww links only adjacent words in an entity. Thus it is straightforward to revert the process to map a \sgraph and an \ergraph by using rules. As a result, an \sgraph is isomorphic to an \ergraph.





\subsection{\ourmodel Model}
\label{sec:gnn_model}
In this section, we present \ourmodel that parses a document image to a \sgraph, which is reformulated as predicting relations between words in a document. In contrast to prior studies~\cite{layoutxlm,layoutlm}, our model relies on relative distances between the bounding boxes of words, which are robust across languages, rather than linguistic features. Moreover, we adopt Graph Attention Network (GAT) \cite{GAT} to capture the similarities of relations between the neighbours of a word. However, we slightly modify GAT model to include edge features in addition to node features in message passing. We do this by concatenating our edge features with node features before performing the message passing. The coherence of predicted relations and graph properties of a \sgraph are ensured by ILP at the inference stage.  


Similar to prior works~\cite{layoutlm,layoutxlm}, we apply the off-the-shelf OCR model \cite{doctr2021} to map document images to bags of words. The OCR outputs are reorganized by sorting lines from top to bottom and arranging words from left to right in their original order by using their bounding-box coordinates. For each word $w$, we define its neighbourhood $\gN_w$ as the set of words, to which it can potentially have relations. The set $\gN_w$ consists of at most $K$ nearest words, including the word to the right of $w$ and the words below it. This definition of neighbourhood is motivated by the fact that the edge score between two words is independent of the orientation of the edge. Hence, during inference, we only compute edge scores between a word and each word in its neighbourhood $\gN_w$ to avoid repeated computations.

We observe that a significant proportion of the neighbours of a word have the same relations. For example, all words in an answer composed of multiple words are associated with the relation \ww. By viewing a form as a \sgraph, we take the GAT model as our backbone because it supports both node features and edge features. Herein, each node $w$ is represented by a normalized bounding box coefficient vector $\vh_w = (x_{left}/W, y_{top}/H, x_{right}/W, y_{bottom}/H)$, where $W$ and $H$ denotes the width and the height of a document image respectively. To support cross-lingual parsing, we represent an edge $(v_a, v_b)$ only by a relative spacing feature vector $\vd_{ij}$. Such a feature vector $\vd_{ij}$ is computed based on both horizontal and vertical relative distances between their normalized bounding boxes. Specifically, given two words $w_{i}$ and $w_{j}$ with the bounding boxes ($x_{i1}, y_{i1}, x_{i2}, y_{i2}$), and ($x_{j1}, y_{j1}, x_{j2}, y_{j2}$) respectively, the spacing between $w_{i}$ and $w_{j}$ is calculated along x and y-axis by 
\begin{displaymath}
    d_{x1}, d_{x2}, d_{x3} = (x_{i1} - x_{j1}), (x_{i2} - x_{j1}), (x_{i2} - x_{j12})
\end{displaymath}
\begin{displaymath}
    d_{y1}, d_{y2}, d_{y3} = (y_{i1} - y_{j1}), (y_{i2} - y_{j1}), (y_{i2} - y_{j12})
\end{displaymath}
As a result, the spacing feature vector $\vd_{ij} = [d_{x1}, d_{x2}, d_{x3}, d_{y1}, d_{y2}, d_{y3}]$.

In comparison with the prior works~\cite{layoutlm,layoutlmv2,layoutlmv3,layoutxlm}, which use text, image, and layout features altogether, our model is more parameter efficient by adopting only language-independent and layout-agnostic features. Due to those simple features, our model contains only 7.3K parameters, significantly less than those transformer-based models, which have approximately 200-400M parameters.

\subsection{Inference}
\label{sec:inference_sec}
We formulate the inference problem as an integer linear program that aims to identify the most likely \sgraph satisfying the graph properties outlined in Sec. \ref{sec:word-relation_graph}.

Given an edge embedding $\ve_{ij}$, we compute a score $s_{ijz}$ for each possible label $z \in \gZ^w$ by using a linear layer. As a \sgraph is sparse, we extend $\gZ^w$ to the set $\gZ^+$ by adding a relation \nl, which indicates there is no relation between two words. Let $u_{jiz} \in \{0,1\}$ be a binary variable indicating if there is an edge with a label $z$ between $v_i$ and $v_j$, the solution to the following integer linear program is the most likely \sgraph.
\begin{equation}
    \begin{array}{ll@{}ll}
\text{min}  & \displaystyle \vs^{\text{T}} \vu\\ 
\text{s.t.}& \displaystyle \rmA \vu \leq \vb
\end{array}
\end{equation}
where $\rmA$ denotes the constraint matrix and $\vb$ is the corresponding constant vector.

We formulate five constraints specific to documents for efficient joint inference using integer linear programming:
\begin{itemize}
\item \textbf{Connectivity constraint (C1):} The generated \sgraphs are fully connected such that there is a path between any pair of nodes in a graph. 
\begin{displaymath}
    \sum_{v_j \in \gN_i}\sum_{z \in \gZ^r} u_{ijz} \geq 1, \forall v_i \in \gV^w
\end{displaymath}
\item \textbf{QA constraint (C2):} If a relation between two words $u_{ijz^q}$ is \qa, where $z^q$ denotes \qa, then the next immediate relation must be either \proximateh ($z^h$) or \ww ($z^w$). Let the next relation is denoted by $u_{i(j+1)z}$, this constraint is defined as:   

\begin{displaymath}
   u_{i(j+1)z^h} + u_{i(j+1)z^w} \geq u_{ijz^q}
\end{displaymath}

\item \textbf{Single label constraint (C3):} There is only one relation in $\gZ^w$ between any pair of nodes in a \sgraph. 

\begin{displaymath}
    \sum_{z \in \gZ^+} u_{jiz} = 1, \forall (v_i, v_j) \in \gV^w \times \gV^w
\end{displaymath}

\item \textbf{At least one semantic relations (C4):} A word is part of an entity. Therefore, it is linked to another word via \ww if the entity contains multiple words. If the word is at the beginning or the end of an entity, it points to another entity via either \qa or \hq. In other words, a word is involved in at least one relation in $\gZ^s = \{\qa, \ww, \hq\}$. If an entity contains a single word, it should be associated with either \qa or \hq. Otherwise, if a word is associated only with $proximate_V$ or $proximate_H$, we cannot determine which entity it belongs to and the type of the corresponding entity. 

\begin{displaymath}
    \sum_{j \in \gN} \sum_{z \in \gZ^s} u_{jiz} \geq 1, \forall v_i
\end{displaymath}

\item \textbf{At least one semantic relation in the neighbourhood (C5):} The previous constraint can still fail to exclude the cases that a word is only involved in \ww in $\gZ^s$. In such a case, the word is expected to be at either end of an entity. If we need to infer the type of entity, this word should be linked to another word via either \qa or \hq. Therefore, if a word is linked with a \ww relation, it should be linked to a different word with a relation in $\gZ^s$. 

\begin{displaymath}
 u_{i(i+1)z^h} + u_{i(i+1)z^w} + u_{i(i+1)z^q} \geq u_{(i-1)iz^w}
\end{displaymath}

\end{itemize}
After parsing a document image into a \sgraph, we apply the deterministic rules introduced in Sec. \ref{sec:word-relation_graph} to convert the graph into an \ergraph.

\subsection{Model Training}
\label{sec:training}
Inspired by~\cite{domke2013structured}, we apply the cross-entropy loss to each pair of nodes during training. The construction of \sgraphs is realized by the ILP-based inference method detailed above. 
\begin{equation}
    \mathcal{L} = -\sum_{ij}\sum_{z \in \gZ^+} z_{ij} \log{\hat{z_{ij}}}
\end{equation}
where $z_{ij}$ and $\hat{z_{ij}}$ denote the ground-truth labels and predicted labels respectively.


\section{Experiments}
\label{sec:experiments}
We compare our model with the state-of-the-art methods on both monolingual and multilingual form datasets. The results show that our models significantly outperform the baselines in terms of relation prediction on \sgraphs in both settings, which leads further to superior performance on \ergraphs. The extensive ablation studies demonstrate the effectiveness of incorporating structural information using GAT and the constraints during ILP-based inference.  

\subsection{Datasets}
\textbf{FUNSD} \cite{funsd} It is a form understanding dataset consisting of 199 noisy documents which are fully annotated. It has a total of 9,707 semantic entities over 31,485 words. In the official data split, the 199 samples are divided into 149 training samples and 50 testing samples. However, we further split the 149 samples into 139 training and 10 validation samples. Our test split is the same as the official split. Each entity is labelled with one of the four semantic entity labels - "question", "answer", "header", and "other". This dataset is widely employed for semantic entity labelling and relation extraction tasks. 

\noindent\textbf{XFUND} \cite{xfund} This is a multilingual benchmark dataset comprising 199 forms that are labeled by humans in 7 languages, which are Chinese (ZH), Japanese (JA), Spanish (ES), French (FR), Italian (IT), German (DE), Portuguese (PT). The dataset is divided into 149 forms for training and 50 for testing. In the ground-truth annotations, each entity is labeled with either "question", "answer", "header", or "other". Following previous works \cite{layoutxlm,lilt}, we use this dataset to compare our model with existing state-of-the-art models in the zero-shot settings.

To obtain the relation annotations in \ergraphs on both datasets, we apply rules to generate the relation annotations based on entity annotations, followed by manually checking all document images for correctness. Due to the isomorphism between \sgraphs and \ergraphs, we map the resulting \ergraphs to \sgraphs by using the deterministic rules. 

\begin{table}[htb]
\centering
\begin{tabular}{l|ccccc}\toprule
Model & $\#$Parameters & Modality & Precision ($\uparrow$) & Recall ($\uparrow$) & F1 ($\uparrow$) \\\hline
XLM-RoBERTa$_{BASE}$ &  - & T & 0.563 & 0.561 & 0.561\\
InfoXLM$_{BASE}$ &  - & T & 0.593 & 0.603 & 0.598\\
LayoutLM &  11M & T+L+I & 0.664 & 0.666 & 0.665 \\
LayoutXLM$_{BASE}$ &  30M & T+L+I & 0.709 & 0.715 & 0.712\\ 
LayoutLMv2 &  11M & T+L+I & 0.722 & 0.712 & 0.717 \\
StructuralLM &  11M & T+L & 0.741 & 0.749 & 0.745 \\
LiLT[InfoXLM]$_{BASE}$ &  11M & T+L & 0.792 & 0.786 & 0.789\\
LayoutLMv3 &  11M & T+L+I & 0.801 & 0.809 & 0.805 \\
\textbf{Ours} &  8.1K & L & 0.837 & 0.854 & 0.845\\
\textbf{Ours + Constraints} &  8.1K & L & \textbf{0.848} & \textbf{0.861} & \textbf{0.854}\\ \bottomrule
\end{tabular}
\caption{Relation extraction (RE) results on the \sgraphs of FUNSD dataset, where T, L, I denotes if models use text, layout, and image features respectively. Ours denotes \ourmodel without applying ILP-based inference. M stands for million and K refers to thousand. T, L, and I refer to text, layout, and image modalities respectively.}
\label{tab:main}
\end{table}

\begin{table}[htb]
\centering
\begin{tabular}{l|ccc}\toprule
Model & Precision ($\uparrow$) & Recall ($\uparrow$) & F1 ($\uparrow$) \\\hline
LayoutLM & 0.753 & 0.757 & 0.755 \\
LayoutXLM$_{BASE}$ & 0.791 & 0.796 & 0.793\\ 
LayoutLMv2 & 0.847 & 0.852 & 0.849 \\
LiLT[InfoXLM]$_{BASE}$ & 0.864 & 0.881 & 0.872\\
LayoutLMv3 & 0.898 & 0.903 & 0.900 \\
\textbf{\ourmodel} &  \textbf{0.921} & \textbf{0.936} & \textbf{0.928}\\ \bottomrule
\end{tabular}
\caption{Evaluation of entity recognition on the \ergraphs of the FUNSD dataset.}
\label{tab:entity-relation}
\end{table}

\subsection{Implementation Details}
The \ergraphs are constructed using Deep Graph Library (DGL). We use a single-layer GAT with 3 heads and a hidden dimension size of 64 for both node and edge features. We train our model using Adam optimizer for 500 iterations with a learning rate of $1\times e^{-3}$. If the performance on the validation data does not improve after 100 iterations, training stops early. During training, we save the model checkpoint based on its performance on validation data. For inference on the test set, the checkpoint with the best performance across all training epochs is loaded. We train our model on 1 GTX 1080Ti 12 GB GPU.

\subsection{Monolingual Results on FUNSD}

We first evaluate our model on the \sgraphs on FUNSD by considering the task as relation extraction between words. More specifically, we run the state-of-the-art models LayoutLM \cite{layoutlm}, LayoutLMv2 \cite{layoutlmv2}, LayoutLMv3 \cite{layoutlmv3}, StructuralLM \cite{structurallm}, LayoutXLM \cite{layoutxlm}, InfoXLM$_{BASE}$ \cite{infoxlm}, and LiLT[InfoXLM]$_{BASE}$ \cite{lilt} to predict entities, followed by relation extraction. 

For baselines, we perform relation extraction by following the approach in \cite{layoutxlm}. First, we create all possible entity pairs as relation candidates. Each candidate is represented by the concatenation of the corresponding entity representations. Furthermore, the representation of an entity is first constructed by concatenating the embedding of the first token of each entity and the entity type embedding, followed by feeding them through two position-wise feed-forward networks (FFN) modules. The resulting relation candidate representations are fed into a bi-affine classifier for relation classification. The conversion from \ergraphs to \sgraphs is performed by the same set of deterministic rules introduced in Sec. \ref{sec:word-relation_graph}.


Table \ref{tab:main} reports the relation extraction results on \sgraphs in terms of {\it Precision}, {\it Recall}, and {\it F1}. For each metric, we take the micro-average among the relations in $\gZ^w$. The two variations of our model achieve superior performance over the baselines by only using the relative spacing features. In contrast, all baselines use textual features extracted from large language models that require pre-training on large-scale datasets. Some of the baselines, such as variations of LayoutLM, require even vision features. The number of parameters of our models is also significantly smaller than their competitors. Although pre-trained language models are widely used in a number of AI applications, our work raises the basic question for future research ``Are language models necessary for form recognition?''. 

Apart from relation extraction, we also evaluate the models in terms of entity recognition on converted \ergraphs. Table \ref{tab:entity-relation} summarizes the results based on an exact match in terms of Precision, Recall, and F1.

\begin{table*}[t]
\centering
\scalebox{0.95}
{
\begin{tabular}{l|cc|c|ccccccc|c}\toprule
\multirow{2}{*}{Model} &\multicolumn{2}{c|}{Pretraining} & \multicolumn{1}{c|}{FUNSD} & \multicolumn{7}{c|}{XFUND} & \multirow{2}{*}{Avg} \\\cline{2-11}
& Language & Size &EN & JA & ZH & DE & FR & PT & ES & IT\\\hline
XLM-RoBERTa$_{BASE}$ &  - & - & 0.587 & 0.116 & 0.132 & 0.335 & 0.400 & 0.354 & 0.281 & 0.286 & 0.311\\
InfoXLM$_{BASE}$ &  - & - & 0.601 & 0.132 & 0.159 & 0.357 & 0.398 & 0.352 & 0.295 & 0.300 & 0.324\\
LayoutXLM$_{BASE}$ &  Multilingual & 30M & 0.701 & 0.258 & 0.240 & 0.443 & 0.568 & 0.549 & 0.461 & 0.499 & 0.464\\
LiLT[InfoXLM]$_{BASE}$ &  English & 11M & 0.771 & 0.303 & 0.349 & 0.562 & \textbf{0.691} & 0.613 & 0.554 & 0.586 & 0.553\\
\textbf{Ours} &  $\times$ & - & 0.845 & 0.611 & 0.625 & 0.636 & 0.679 & 0.641 & \textbf{0.651} & 0.625 & 0.664\\
\textbf{Ours + Constraints} &  $\times$ & - & \textbf{0.853} & \textbf{0.625} & \textbf{0.626} & \textbf{0.647} & 0.673 & \textbf{0.641} & 0.644 & \textbf{0.638} & \textbf{0.669}  \\ \bottomrule
\end{tabular}
}
\caption{Multilingual relation extraction (RE) results on \sgraphs of XFUND dataset.}
\label{tab:zero-shot}
\end{table*}

\subsection{Zero-Shot Multilingual Results}

We further evaluate model performance by applying the models trained on FUNSD directly to the scanned forms in other languages on XFUND. Herein, we compare our models with the state-of-the-art methods: XLM-RoBERTa$_{BASE}$, InfoXLM$_{BASE}$, LayoutXLM$_{BASE}$ \cite{layoutxlm}, and LiLT[InfoXLM]$_{BASE}$ \cite{lilt}. The latter two models are pre-trained on large-scale document datasets of size 30M and 11M respectively.

Table~\ref{tab:zero-shot} reports the corresponding zero-shot multilingual relation extraction results on the \sgraphs of XFUND dataset. Overall, our best model outperforms the baselines in 7 out of 8 languages in terms of \textit{F1}. The geometric mean of the F1 is more than 10\% better than the strongest baseline. It is noteworthy that performance improvement is achieved without any time-consuming pre-training. We only use the 139 forms from the FUNSD dataset to train our models. Our models based on relative spacing features are more transferrable than those multilingual language models on this task. A further inspection shows that our models benefit from the fact that the space between two words within an entity, as well as the distance between a question word and an answer word, are similar across languages. In this zero-shot multilingual setting, incorporating the constraints into inference does not always help, though it leads to improvements in 6 out of 8 languages. 


\begin{table}[t]
\centering
\begin{tabular}{l|c|ccc}\toprule
Model & Precision & Recall & F1\\\hline
GraphSAGE &  0.756 & 0.768 & 0.761\\
GCNs &  0.814 & 0.823 & 0.818\\
\ourmodel - Edge$_{feats}$ &  0.705 & 0.717 & 0.710\\
\ourmodel + Edge$_{feats}$ &  0.837 & 0.854 & 0.845\\
\ourmodel + Edge$_{feats}$ + all constrs &  \textbf{0.848} & \textbf{0.861} & \textbf{0.854}\\
\ourmodel + Edge$_{feats}$ + all constrs - C4+C5 &  0.846 & 0.860 & 0.852\\
\ourmodel + Edge$_{feats}$ + all constrs - C1 &  0.845 & 0.858 & 0.851\\
\ourmodel + Edge$_{feats}$ + all constrs - C2 &  0.840 & 0.851 & 0.845\\
\bottomrule
\end{tabular}
\caption{Ablation studies. C1, C2, C4 and C5 are the constraints defined in Sec \ref{sec:inference_sec}}
\label{tab:ablation}
\end{table}

\begin{table}[t]
\centering
\begin{tabular}{c|c}\toprule
Constraints & Number of violations\\\hline
C1 &  71\\
C2 &  169\\
C3 &  0\\
C4 &  51\\
C5 &  10\\
\bottomrule
\end{tabular}
\caption{This table illustrates the number of predictions by \ourmodel that violated a constraint. These are corrected by applying ILP inference described in Sec \ref{sec:inference_sec}.}
\label{tab:constraints_violations}
\end{table}


\begin{table}[t]
\centering
\begin{tabular}{l|c|ccc}\toprule
Model & Precision & Recall & F1\\\hline
\ourmodel + Edge$_{feats}$ in lower-level and at classifier &  \textbf{0.837} & \textbf{0.854} & \textbf{0.845}\\
\ourmodel + Edge$_{feats}$ in lower-level and not at classifier &  0.713 & 0.725 & 0.718\\
\ourmodel + Edge$_{feats}$ not in lower-level but at classifier &  0.831 & 0.849 & 0.840\\
\ourmodel - Edge$_{feats}$ &  0.705 & 0.717 & 0.710\\
\bottomrule
\end{tabular}
\caption{Effectiveness of edge features.}
\label{tab:node_edge_feats}
\end{table}

\subsection{Ablation Study}
\begin{figure}[t]
    \begin{center}
    \includegraphics[width=0.80\textwidth]{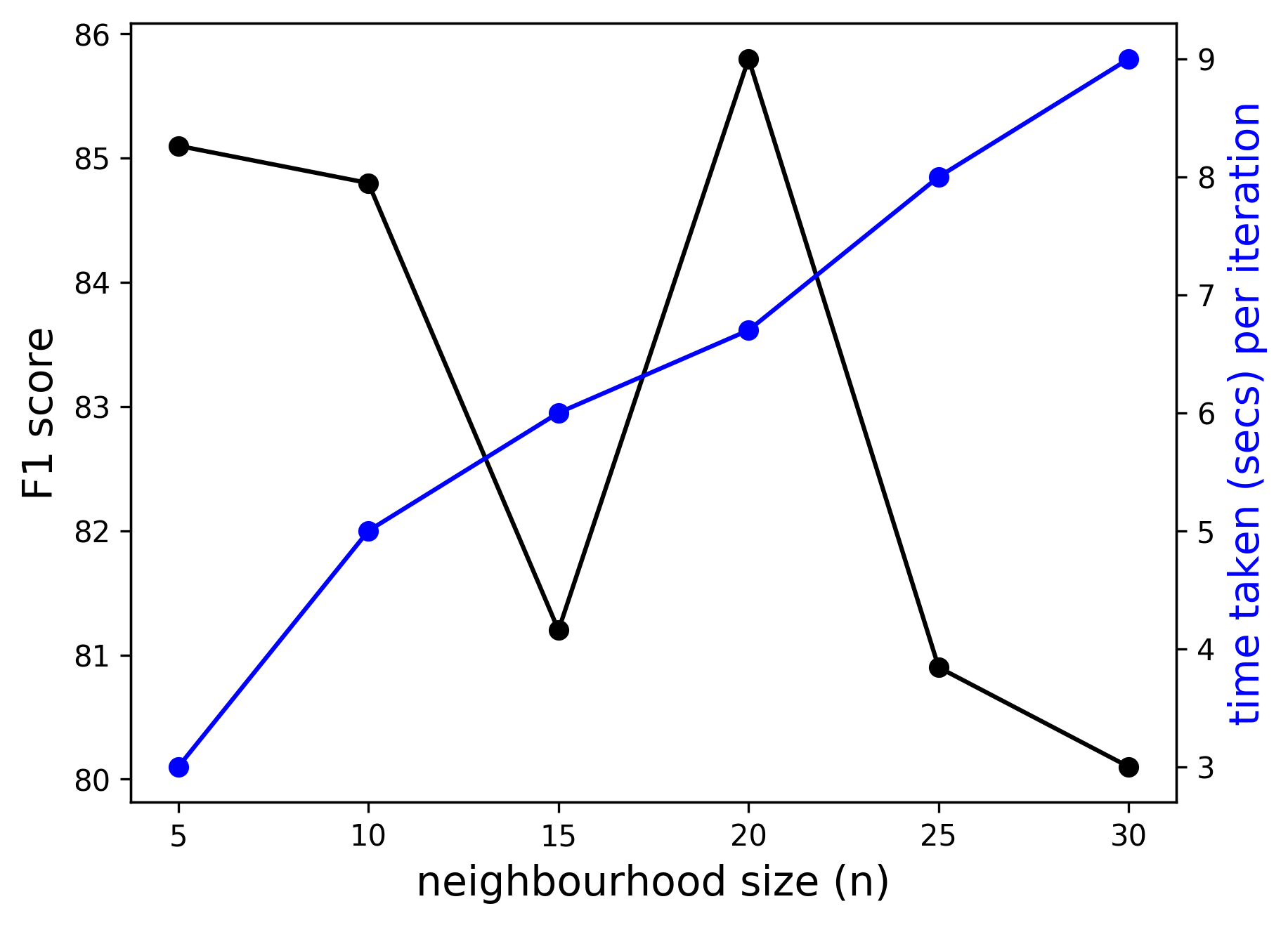}
    \end{center}
    \caption{Tradeoff between neighbourhood size and computational time.}
    \label{fig:tradeoff}
\end{figure}
We evaluate the effectiveness of using GAT, edge features, constraints in ILP, and the size of neighbourhood in our model. To show the usefulness of GAT, we compare it with GraphSAGE \cite{hamilton2017inductive} and graph convolutional networks (GCNs) \cite{gcn} by using the same features. For edge features, we run ablation studies by removing them with or without applying ILP. To understand the usefulness of the constraints during inference, we remove the connectivity constraint (C1), QA constraint (C2), and the last two semantic constraints (C4+C5) respectively from the full model. 

As shown in Table~\ref{tab:ablation}, it is expected that our full model performs the best among all variations. Removing relative spacing features leads to the largest drop in terms of all metrics. GAT demonstrates its strengths over the two alternative neural structures. Removing any of the constraints, the model performance drops slightly. Among them, C2 is clearly the most useful one among them in terms of improving performance. As such, C2 is the qa constraint that is designed to correct the wrongly predicted \qa relation labels, which are one of the most frequent model prediction errors based on our evaluation. Before applying ILP, we observe that there were 169 out of 51775 word pairs in the test set that violate this constraint. Table \ref{tab:constraints_violations} illustrates the number of violations in the predictions of \ourmodel before applying the constraint-based inference.

The constraints C4 and C5 are designed to ensure isomorphism between \ergraphs and \sgraphs. This constraint assists in preventing any isolated word pairs that are not connected to an entity (question, or header). For instance, if a question has more than one word in the answer, the words within the answer are linked together using the \ww  relation. The last word of the question and the first word of the answer are linked by the \qa relation. When transforming a \sgraph to an \ergraph, we take into account the rule that any word pairs with a \ww connection following a \qa relation are mapped to the entity relation "answer" in the \ergraph. However, if \ourmodel predicts a different relation for a word pair within the answer than \ww, this will result in an inaccurate mapping of \sgraph to \ergraph. Additionally, a word pair can only have a \ww relation if it is part of a chain of words whose head is linked to either a "question" or a "header." It is impossible to have an isolated word pair with a \ww relation that isn't related to either the "question" or "header" in our proposed way of \sgraph construction.  However, a few of the \ourmodel model's relation predictions could result in isolated word pairs. In order to prevent this, our uniquely designed semantic constraints correct these mistakes and aid in the accurate mapping of the \sgraph to the \ergraph. There are 134 total isolated word pairings before semantic constraints are applied. However, with the application of the constraints, there were no violations, and the performance benefit from using semantic constraints—which are made particularly to achieve isomorphism between \sgraph and \ergraph—is very little.\\\\
\noindent\textbf{Where to put edge features} To demonstrate the significance of using spacing between pairs of nodes as its edge feature, we conduct four experiments with and without edge features. The results for these experiments are available in Table \ref{tab:node_edge_feats}. Our key objective is to determine whether it is more effective to concatenate edge features with node features when computing the node attention rather than doing so only at the model's final classifier layer. It is evident that employing edge features solely while computing node attention (row 2) results in much lower performance than using the model's final (classifier) layer (row 3). Nevertheless, it is still marginally preferable to not use edge features at all (row 4). The optimal performance can be obtained by employing edge features while computing node attention as well as at the last layer of the model (row 1). \\\\

\noindent \textbf{Effect of neighbourhood size} On both datasets, the "linking" key found in the JSON annotation files of the ground-truth documents is used to construct the ground-truth graphs. There are no ground-truth annotation files accessible during the inference. Therefore, the model must decide which nodes should be connected to which nodes. The "no-link" relation label enables the model to discover which nodes ought to be connected by an edge. Each word or node in the graph is linked to the $n$ following words or nodes in the document using an edge labelled \nl. The edges predicted with a "no-link" relation label are removed prior to applying the ILP inference. We experiment with various $n$ to examine the impact of neighbourhood size on the model's performance. As $n$ increases, the computational time increases. We determine an ideal value for $n$ by finding an optimal trade-off between computation time and performance.

\section{Conclusion}
\label{sec:conclusion}
In this work, we propose a novel \ergraph parsing model called \ourmodel that is language independent. Our model parses a document image into a \sgraph in order to minimize error propagation of pipeline approaches and reduce the time complexity of inference. This graph is then transformed by deterministic rules into a fully linked \ergraph. Due to this simplification, inference time is sharply reduced. Our model simply takes into account relative spacing features extracted from layout information in order to allow language-independent form understanding without the use of pre-trained language models. To ensure that the generated graphs match the specifications of \ergraphs, we use ILP-based inference by incorporating designated constraints for this task. Our experimental results on the multilingual XFUND and FUSND datasets demonstrate that our proposed approach produces superior results over competitive baselines. In particular, on the zero-short multilingual form understanding task, our model surpasses recent strong baselines by a large margin. Additionally, we conduct extensive ablation studies that demonstrate the effectiveness of each new design choice we proposed.

%
%
%
\bibliographystyle{splncs04}
\bibliography{mybibliography}
\end{document}